# Linguistically Informed Graph Model and Semantic Contrastive Learning for Korean Short Text Classification

JaeGeon Yoo[1], Byoungwook Kim[2], Yeongwook Yang[2(✉)], and Hong-Jun Jang[1(✉)]

[1] Dept. of Data Science, Kangwon National University, Chuncheon 24341, Republic of Korea
{yoojg,hongjunjang}@kangwon.ac.kr
[2] Dept. of Computer Science and Engineering, Gangneung-Wonju National University, Wonju 26403, Republic of Korea
{bwkim,yeongwook.yang}@gwnu.ac.kr

**Abstract.** Short text classification (STC) remains a challenging task due to the scarcity of contextual information and labeled data. However, existing approaches have pre-dominantly focused on English because most benchmark datasets for the STC are primarily available in English. Consequently, existing methods seldom incorporate the linguistic and structural characteristics of Korean, such as its agglutinative morphology and flexible word order. To address these limitations, we propose LIGRAM, a hierarchical heterogeneous graph model for Korean short-text classification. The proposed model constructs subgraphs at the morpheme, part-of-speech, and named-entity levels and hierarchically integrates them to compensate for the limited contextual information in short texts while precisely capturing the grammatical and semantic dependencies inherent in Korean. In addition, we apply Semantics-aware Contrastive Learning (SemCon) to reflect semantic similarity across documents, enabling the model to establish clearer decision boundaries even in short texts where class distinctions are often ambiguous. We evaluate LIGRAM on four Korean short-text datasets, where it consistently outperforms existing baseline models. These outcomes validate that integrating language-specific graph representations with SemCon provides an effective solution for short text classification in agglutinative languages such as Korean.

## 1    Introduction

Short text classification (STC) is a natural language processing (NLP) task that automatically assigns short sentences such as search queries, social media posts, or news headlines to predefined labels [1]. It is essential for applications including information retrieval, sentiment analysis, news recommendation, and intent detection, and its importance has increased with the rapid growth of short-text data [2-4], particularly in personalized or real-time services where classification quality directly affects service performance.

However, short texts contain limited context and often exhibit irregular or incomplete structures, making semantic interpretation difficult [5]. For example, the Korean sentence "병원 갔다"(went to the hospital) may indicate going to receive medical treatment, to visit a patient, to undergo a medical examination, or to work there as medical staff, depending on omitted particles and context, while "배터리 갈다"(change the battery) may express either a command or a statement about replacing the battery. Such ambiguity arises because Korean short sentences frequently omit postpositions and endings, resulting in fewer contextual cues and higher semantic uncertainty than in fixed word-order languages. In practice, this issue is further aggravated by label scarcity and data imbalance. Consequently, limited length, incomplete structure, and label sparsity jointly degrade STC performance [6,7].

Various approaches have been proposed to address these challenges. Early methods used statistical features such as Bag-of-Words (BoW) [8] and TF-IDF [9] with classifiers like SVMs [10]. Deep neural models including CNNs [11] and LSTMs [12] later enabled end-to-end learning but still struggled to capture latent semantics in short sentences. More recently, graph-based methods represent words, part-of-speech tags, and named entities as nodes and learn document relationships through GNNs [13], showing promising ability to compensate for contextual sparsity.

Nevertheless, a review of the existing literature indicates that studies on Korean STC remain limited, particularly those that explicitly model Korean linguistic structures. Existing studies largely rely on general-purpose models that overlook fine-grained linguistic structure. Korean is an agglutinative language where meaning is composed at the morpheme level and conveyed through postpositions and endings [14], and sentence interpretation can vary with their presence or word-order changes [15]. These properties differ fundamentally from English and become more critical in short texts where such markers are often omitted, causing semantic loss. As a result, English-centric models frequently produce distorted interpretations, since prior work has rarely addressed how to reconstruct missing context through relational modeling. Effective Korean STC therefore requires architectures that explicitly encode and restore language-specific structural cues across multiple levels [16].

Motivated by this gap, we propose LIGRAM (Linguistically Informed Graph Model), a hierarchical heterogeneous graph framework for Korean STC. LIGRAM constructs morpheme, part-of-speech (POS), and named-entity subgraphs to explicitly model hidden linguistic structure. The morpheme graph alleviates whitespace-tokenization limitations in agglutinative languages, the POS graph models grammatical relations to compensate for omitted particles, and the entity graph provides semantic anchors for disambiguation. Their hierarchical integration enables modeling beyond surface features and captures Korean-specific linguistic constraints.

To further improve semantic discriminability, we assume that documents in the same class share latent thematic structure. We therefore represent each document as a topic distribution reflecting its semantic cluster membership, where each dimension indicates the probability of belonging to a cluster.

Finally, we apply semantics-aware contrastive learning based on similarity between topic distributions [17]. Unlike instance-level contrastive learning [18], which may push apart semantically related sentences due to surface differences, our method forms positive pairs according to topical meaning, encouraging representations aligned with true decision boundaries.

Experimental results on multiple Korean short-text datasets demonstrate that the proposed LIGRAM model consistently outperforms baseline methods, verifying the effectiveness of language-specific graph integration and semantics-aware representation learning for Korean short-text classification. The main contributions of this paper are as follows:

- We propose LIGRAM, a hierarchical heterogeneous graph model that integrates Korean-specific linguistic units (morphemes, POS tags, named entities) to capture structural cues for short-text classification.
- We present SemCon, a semantic-aware contrastive learning that forms contrastive pairs based on pseudo-topic distributions for clearer class separation.
- We conduct experiments on Korean short text classification datasets and show that our method outperforms existing graph-based and contrastive learning baselines.

## 2    Related Work

### 2.1    Graph-based Short Text Classification

STC remains challenging due to limited context and incomplete syntactic structures. Graph neural networks (GNNs) have recently emerged as an effective solution by explicitly modeling relationships between words and documents and incorporating structural linguistic information.

GNN-based methods are generally categorized into two types. The first constructs a graph for each document, representing words as nodes and co-occurrence as edges; representative models include TLGNN [19], TextING [20], and HyperGAT [21]. However, their performance often degrades when labeled document graphs are scarce. The second builds a single heterogeneous corpus-level graph containing both document and word nodes and predicts labels for document nodes. TextGCN [3] exemplifies this approach and performs well in semi-supervised settings with few labels, but its effectiveness is limited in short texts due to sparse word connections.

To mitigate these issues, later studies incorporate richer linguistic structure into GNNs. HGAT [6] jointly models topics, entities, and documents with attention, and STGCN [13] integrates document–word–topic graphs with a BiLSTM encoder for improved contextual representation. SHINE [22] and GIFT [23] further adopt hierarchical heterogeneous graphs that combine words, POS tags, and named entities to enhance information propagation.

Despite these advances, most graph-based STC models are designed for English-centric corpora and do not fully reflect the linguistic properties of agglutinative languages. Korean shows strong morpheme-level semantic variation, encodes grammatical roles through postpositions and endings, and permits flexible word order, making sentence meaning difficult to capture with word-level links alone, especially in short texts with minimal context. Prior Korean STC studies have largely relied on statistical features or basic morphological analysis, which struggle to model the structural relations needed to recover missing context.

Accordingly, effective Korean STC requires language-specific graph representations that integrate multiple linguistic cues, including morphemes, POS tags, and named entities, within a unified framework. In line with this need, our study designs a hierarchical heterogeneous graph model that explicitly captures Korean grammatical dependencies and semantic composition to better reflect the language's structural constraints.

## 2.2    Contrastive Representation Learning

Contrastive Learning (CL) is a self-supervised approach for learning semantically meaningful representations from unlabeled data [24,25] and has shown strong performance across computer vision, NLP, and graph learning. Its objective is to pull semantically similar samples closer while pushing dissimilar ones apart to form a discriminative representation space [26]. Early instance-discrimination methods treated each sample as a separate class and ignored inter-sample semantic similarity, limiting their ability to capture higher-level semantics [27].

In NLP, CL commonly constructs positive pairs via data augmentation, but short texts provide limited information, making performance sensitive to augmentation quality. Later work improves this paradigm by incorporating label supervision [28,29] or prototype guidance [30]. However, ambiguous topic boundaries in short texts can still cause semantically related sentences to be separated in the embedding space, degrading classification accuracy.

To alleviate this issue, we adopt the topic-distribution-based contrastive framework of TSCTM [17], where document embeddings are converted into pseudo-topic distributions and contrastive learning is performed according to topical similarity, allowing the model to learn representations with clearer semantic boundaries and stronger class separation.

## 3    Method

This study proposes a hierarchical heterogeneous graph neural network architecture designed to address both the lack of contextual information in Korean short-text classification and the linguistic and structural characteristics of the Korean language. To this end, a set of heterogeneous subgraphs is constructed according to different linguistic cues. Each subgraph is formally defined as

$$G_\pi = (V_\pi, X_\pi, A_\pi), \pi \in \{w, p, e\} \tag{1}$$

where $V_\pi$ denotes the node set, $X_\pi$ represents the feature matrix containing node embeddings, and $A_\pi$ is the adjacency matrix describing the relational structure between nodes. Specifically, $\pi$ corresponds to the morpheme ($w$), part-of-speech ($p$), and entity ($e$) subgraphs, which collectively encode diverse linguistic cues of Korean. The proposed model first constructs these subgraphs independently and then integrates them hierarchically to learn comprehensive document representations. In addition, a semantics-aware contrastive learning strategy is applied to align semantically similar document embeddings and to reinforce clear semantic boundaries within the latent space. The overall architecture of the proposed model is illustrated in Fig. 1.

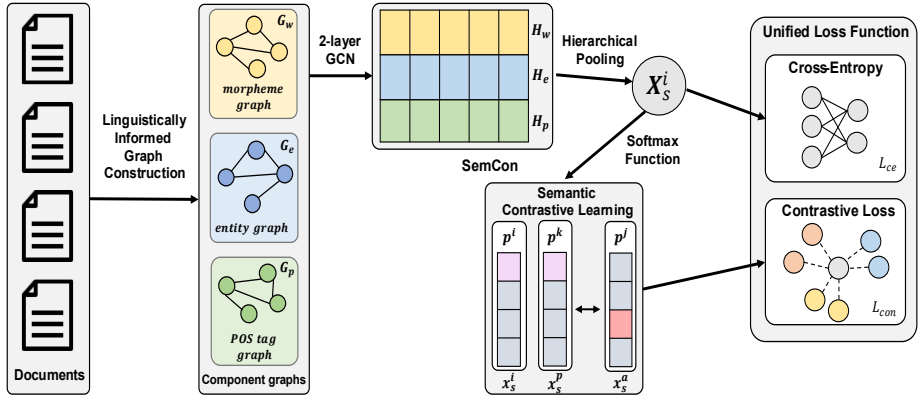

**Fig. 1.** An overall architecture of the proposed model. Based on morphemes ($G_w$), part-of-speech ($G_p$), and entity ($G_e$) three heterogeneous graphs are constructed. A GCN is applied to each graph to obtain document embeddings $x_s^i$. Next, a Softmax function is applied to the document embedding $x_s^i$ to obtain a pseudo-topic distribution $p^i$. Embeddings with the same pseudo-topic serve as positives ($x_s^p$), while those with different pseudo-topics serve as negatives ($x_s^a$). These are used to compute the Semantics-aware Contrastive loss $L_{con}$. The final training objective combines the standard classification loss $L_{ce}$ with the contrastive loss $L_{con}$.

## 3.1    Data Preprocessing

To ensure data quality and effective model training, we perform a series of preprocessing steps on the raw Korean short texts. Specifically, conventional whitespace-based tokenization is inadequate for representing proper semantic units in Korean because content and functional morphemes are tightly integrated within a single word unit (eojeol). To address these challenges, we first perform preprocessing, removing special characters, deduplicating documents, and filtering low-frequency morphemes (frequency < 5).

## 3.2    Graph Construction

**Morpheme Graph ($G_w$):** Korean is an agglutinative language where grammatical meanings are expressed through postpositions and endings attached to content words.

Unlike English, Korean exhibits flexible word order and inconsistent word-level meanings; these challenges are exacerbated in short texts where frequent particle omissions make it difficult to accurately capture syntactic and semantic relationships. To address these linguistic challenges, we construct a morpheme-level graph ($G_w$) by decomposing sentences into morphemes using the Kiwi morphological analyzer, which is effective in disambiguation and sentence boundary recognition. Each node is initialized with a d-dimensional embedding(d=768), $X_w \in \mathbb{R}^{|V_w| \times d}$ from the KLUE/RoBERTa model pretrained on large-scale Korean corpora. The adjacency matrix $A_w$ is defined by Pointwise Mutual Information (PMI) scores between co-occurring morphemes, enabling the model to encode semantic proximity among morphologically related words.

**POS Graph ($G_p$):** In addition to capturing contextual information at the word level, modeling grammatical relations between linguistic components within a sentence plays an important role in short text classification. While the original English-based SHINE model also incorporated POS information, it is even more crucial to explicitly represent it in Korean, where grammatical features such as particles and endings are encoded at the morpheme level rather than the word level. In Korean short texts, these grammatical markers are frequently omitted or modified, which often leads to significant contextual loss and ambiguity. The POS graph ($G_p$) addresses this issue by representing grammatical relationships as explicit nodes, thereby compensating for the missing context through structural modeling. In this study, POS tagging is performed using the Kiwi morphological analyzer, and all POS tags appearing across the corpus are represented as node set $V_p$. Each POS node is encoded as a one-hot vector $X_p$ corresponding to its unique index, and the adjacency matrix $A_p$ is constructed based on the PMI between co-occurring POS tag pairs within the same document. This structure effectively reflects morphological and syntactic variations such as the presence or omission of particles and endings thus improving the precision of contextual understanding in Korean short text classification.

**Entity Graph ($G_e$):** In short texts, named entities such as locations, organizations, and person names often serve as key semantic cues that reveal the document's topic. This role becomes even more crucial in sentences with limited contextual information, where entity mentions provide strong signals for semantic disambiguation. To capture these characteristics, named entities are extracted using the domain-specific KPF-BERT-NER model, which is fine-tuned on the BIO tagging scheme based on the Ko-BERT architecture. Each recognized entity token's final hidden representations are mean-pooled to form the entity node embeddings $X_e$. The adjacency matrix $A_e$ is computed based on the cosine similarity between entity vectors, allowing the model to capture semantic relatedness not only among entities within the same document but also across semantically similar documents. This design strengthens entity-level semantic propagation, which plays a critical role in improving contextual understanding in Korean short text classification.

For each of the three subgraphs $G_\pi$ ($\pi \in \{w, p, e\}$), a GCN is applied as follows:

$$H_\pi = \tilde{A}_\pi \cdot ReLu(\tilde{A}_\pi X_\pi W_\pi^1) W_\pi^2 \qquad (2)$$

where $\tilde{A}_\pi = \tilde{D}^{-\frac{1}{2}}\hat{A}\tilde{D}^{-\frac{1}{2}}$ is the normalized adjacency matrix, $\hat{A} = A_\pi + I$ adds self-connections to the original adjacency matrix $A_\pi$, and $\tilde{D}_{ij} = \sum_i [A_\pi]_{ij}$ is the diagonal degree matrix. $W_\pi^1$ and $W_\pi^2$ denote learnable weight parameters, and $H_\pi$ represents the resulting node embeddings for each subgraph. Based on these node embeddings, document-level embeddings are obtained via hierarchical pooling:

$$\hat{x}_\pi^i = u(H_\pi^\top s_\pi^i), \ u(X) = \frac{X}{\|X\|_2} \tag{3}$$

where $s_\pi^i \in \mathbb{R}^{|V_\pi|}$ is an attention vector representing the relevance between document $i$ and nodes in graph $G_\pi$. When $\pi = w, p$ , the attention weights are defined as $[s_\pi^i]_j = TF - IDF(v_j^\pi, s_i)$ , and when $\pi = e, [s_e^i]_j = 1$ if the entity appears in the document, and 0 otherwise. Here, $v_j^\pi \in V_\pi$ denotes the $j$ node in the graph $G_\pi$. The final document embedding is obtained by concatenating the pooled representations from each subgraph $x_s^i = \hat{x}_w^i \| \hat{x}_p^i \| \hat{x}_e^i$. Then, the document-document adjacency matrix is defined as:

$$[A_s]_{ij} = \begin{cases} (x_s^i)^\top x_s^j & if (x_s^i)^\top x_s^j \geq \delta \\ 0 & otherwise \end{cases} \tag{4}$$

Edges are therefore created only when the dot product similarity between document embeddings exceeds a threshold $\delta$.

### 3.3    Contrastive Semantic Representation Learning

To capture semantic boundaries between documents, this study adopts the SemCon framework, a contrastive learning-based representation learning strategy designed to enhance semantic discrimination among short texts. Conventional instance-level contrastive learning often treats semantically similar sentences as negative pairs, which can hinder representation learning in short text classification where topic boundaries are inherently ambiguous.

To address this limitation, we draw inspiration from the semantics-aware contrastive architecture of TSCTM [17], where each document embedding is transformed into a topic distribution through a softmax layer. For each document representation $x_s^i$, a topic probability distribution $p^i = softmax(x_s^i) \in \mathbb{R}^C$ is obtained, where $C$ denotes the number of classes. The class with the highest probability is assigned as the document's pseudo-topic, representing its most likely semantic cluster. Documents sharing the same pseudo-topic are treated as positive pairs, while those with different pseudo-topics are treated as negative pairs. This pseudo-topic mechanism allows the model to learn topic-level semantic relationships without directly relying on gold labels.

The contrastive loss is computed based on the cosine similarity between a document embedding $x_s^i$ and its positive examples $x_s^p$:

$$L_i = -\frac{1}{|P_i|} \sum_{p \in P_i} log \frac{exp \ (sim(x_s^i, x_s^p))}{\sum_{a \in Z_i} exp \ (sim(x_s^i, x_s^a))} \tag{5}$$

$$L_{con} = \frac{1}{N} \sum_{i=1}^{N} L_i \tag{6}$$

where $z_i$ is the set of all documents excluding $i$, and $P_i$ is the set of positive documents that share the same pseudo-topic as document $i$. This objective encourages embeddings of semantically related documents to move closer together, while pushing apart those with different topical meanings. Unlike topic-node-based approaches, the proposed model learns topic-level semantic relations directly from the distribution of document embeddings, without relying on explicit topic structures. This enables the model to acquire semantically consistent representations even for short texts with ambiguous topic boundaries, thereby improving generalization performance under limited-label conditions.

### 3.4    Unified Loss Function

To simultaneously improve classification accuracy and semantic representation quality, this study adopts a unified loss function that combines cross-entropy loss and contrastive loss. The document node $x_s^i$, after being passed through a two-layer GCN, is transformed into a class probability distribution $\hat{y}_s^i$ via the softmax function. The cross-entropy loss is defined as follows:

$$L_{ce} = -\sum_{i \in I_l} y_s^{i^\top} log(\hat{y}_s^i) \tag{7}$$

where $I_l$ denotes the set of indices of labeled documents, and $y_s^i \in \mathbb{R}^C$ is a one-hot vector indicating the ground-truth class of document. The final objective combines the cross-entropy loss and the contrastive loss as follows:

$$L = L_{ce} + \lambda L_{con} \tag{8}$$

where $\lambda$ is a hyperparameter that controls the relative importance of the contrastive loss.

### 3.5    Time complexity

The overall computational complexity of the proposed model consists of three components: (1) construction of heterogeneous subgraphs and two-layer GCN operations, (2) hierarchical pooling, and (3) semantics-aware contrastive learning. For each linguistic subgraph $G_\pi$ ($\pi \in \{w, p, e\}$), the two-layer GCN has a time complexity of $O(E_\pi(d_\pi + d) + 2|V_\pi|d^2)$ where $E_\pi$ and $|V_\pi|$ denote the number of edges and nodes, $d_\pi$ is the input feature dimension, and $d$ is the hidden dimension. By summing across the three subgraphs, the total GCN complexity becomes

$$O\big((E_w + E_p + E_e)(d_\pi + d) + 2(|V_w| + |V_p| + |V_e|)d^2\big) \tag{9}$$

The hierarchical pooling step scales linearly with the number of documents $N$ and active nodes $\bar{m}$, resulting in a complexity of $O(N\bar{m}d)$. Subsequently, constructing the document–document graph based on cosine similarity among document embeddings requires $O(N^2d)$ operations. Finally, the proposed semantics-aware contrastive learning aligns document representations based on their semantic proximity. Since it involves pairwise similarity computation among all documents, the contrastive learning stage also incurs $O(N^2 d)$ complexity. Overall, the total computational complexity of the proposed model can be summarized as

$$O\big((E_w + E_p + E_e)(d_\pi + d) + 2(|V_w| + |V_p| + |V_e|)d^2 + 2N^2 d\big) \qquad (10)$$

where the graph propagation and pairwise contrastive learning are the main computational bottlenecks, but the overall cost remains quadratic in the number of documents, i.e., $O(N^2 d)$, which is relatively simple and stable.

## 4    Experiments

**Table 1.** Dataset Statistics.

| Datasets | # Texts | Avg. length | # Classes | # Train | # Morphemes | # Entities | # POS |
|---|---|---|---|---|---|---|---|
| KLUE YNAT | 14,000 | 7.1 | 7 | 140 | 7,945 | 32,663 | 41 |
| Movie Reviews | 12,739 | 6.3 | 2 | 40 | 2,256 | 8,157 | 45 |
| Snippets | 10,292 | 15.8 | 8 | 160 | 7,372 | 18,334 | 43 |
| Shopping | 12,872 | 6.8 | 2 | 40 | 2,008 | 9,643 | 47 |

### 4.1    Datasets

To evaluate the performance of the proposed model, we utilize four Korean short-text datasets that reflect the characteristics of STC, where sentences are brief and contextual information is limited. The datasets cover both topic classification and sentiment classification tasks. The statistical properties of each dataset are summarized in Table 1.

(1) KLUE YNAT. The YNAT dataset from the KLUE benchmark [31] consists of news headlines collected from Yonhap News Agency between 2016 and 2020, categorized into seven classes: politics, economy, society, culture, world, IT/science, and sports. It reflects the non-standardized and multi-topic characteristics of news headlines, making it suitable for evaluating classification models under ambiguous topic boundaries.

(2) Movie Reviews. The Naver Movie Review dataset is a binary sentiment classification dataset composed of user-generated film reviews. It frequently includes named entities such as actor names, directors, and movie titles, along with colloquial expressions, making it suitable for evaluating representation learning at the morpheme and entity levels.

(3) Snippets (ko) [7]. This dataset consists of web snippets collected from Google search results, translated and refined into Korean, and categorized into eight topics. Each snippet provides a concise summary of a search result, exhibiting grammatical irregularities such as omitted function words and flexible word order, phenomena that are common in Korean short texts.

(4) Shopping. The Naver Shopping Review dataset is a binary sentiment classification dataset comprising short customer reviews on a variety of products including clothing, beauty items, electronics, and food. These reviews primarily focus on aspects such as product quality, delivery speed, and satisfaction compared to expectations, making them effective for evaluating sentiment expression in short contextual texts.

For evaluation, we randomly sample 40 documents per class: 20 are used for training and the remaining 20 for validation, while all other documents are used as the test set.

### 4.2    Baselines

To validate the performance of the proposed model, we compare it with five groups of baseline methods, ranging from traditional feature-based classifiers to advanced graph-based neural models and latest LLMs. Each group differs in the way textual representations are constructed and utilized for classification.

**Group A: Traditional Feature-based Classifiers.**
These methods separate document representation and classification. TF-IDF+SVM and LDA+SVM [10] represent documents as TF-IDF vectors or topic-distribution vectors derived from Latent Dirichlet Allocation (LDA), and perform classification using a linear Support Vector Machine. PTE [5] constructs bipartite graphs among words, documents, and labels to learn word embeddings, which are averaged to form document vectors for classification.

**Group B: Pre-trained Language Model-based Classifiers.**
This group leverages representative Korean pre-trained BERT-based models to obtain contextualized document embeddings, including KLUE/BERT [32], beomi/kcbert-base, and skt/kobert-base-v1. For each model, we compare two strategies: BERT-avg, which averages all token embeddings within a document, and BERT-CLS, which uses the [CLS] token embedding as the document representation. As the three models showed comparable performance, we report their average results.

**Group C: Deep Neural and Graph-based Models.**
Deep learning baselines include CNN and LSTM with two initialization schemes: randomly initialized embeddings (-rand) and pre-trained embeddings (-pre). Graph-based models construct relational structures among words and documents for document-level classification, including TextING, TLGNN, and HyperGAT (document-level graphs), as well as TextGCN and TensorGCN [33], which represent the entire corpus as a single heterogeneous graph.

**Group D: Knowledge-enhanced and Heterogeneous Graph Models.**
STCKA [2] incorporates concept-level knowledge from external sources and applies a BiLSTM-based dual attention mechanism to emphasize informative concepts. HGAT [6] constructs a heterogeneous graph of documents, entities, and topics, and applies hierarchical attention to model inter-node interactions. SHINE integrates linguistic features such as words, POS tags, and named entities into a hierarchical heterogeneous graph to enhance semantic consistency through information propagation.

**Group E: Large Language Models.**
This group encompasses both proprietary API-based models (GPT-5.2, Gemini-2.0-flash) and open-source models (Qwen2.5-7B, Qwen3-4B) to evaluate the performance of state-of-the-art Large Language Models (LLMs). For the API-based models, classification performance was derived through prompt configurations that reflect the same learning structure as the other models. Conversely, the open-source models underwent fine-tuning using Parameter-Efficient Fine-Tuning (PEFT) with LoRA and 4-bit quantization for efficient learning.

**Table 2.** Accuracy and Macro-F1 score on four Korean short-text datasets. The best results are shown in bold excluding LLMs and the second-best results are underlined.

| Group | Model | KLUE YNAT | | Movie Reviews | | Snippets | | Shopping | |
|---|---|---|---|---|---|---|---|---|---|
| | | ACC | F1 | ACC | F1 | ACC | F1 | ACC | F1 |
| (A) | TF-IDF+SVM | 0.3520 | 0.3582 | 0.5540 | 0.5351 | 0.5933 | 0.5853 | 0.6599 | 0.6552 |
| | LDA+SVM | 0.2725 | 0.2657 | 0.5377 | 0.5092 | 0.2795 | 0.2712 | 0.5848 | 0.5698 |
| | PTE | 0.2777 | 0.2781 | 0.5422 | 0.5422 | 0.4426 | 0.4251 | 0.6003 | 0.5975 |
| (B) | BERT-avg | 0.2696 | 0.2694 | 0.5095 | 0.4685 | 0.1982 | 0.1835 | 0.5694 | 0.5224 |
| | BERT-CLS | 0.2587 | 0.2462 | 0.5078 | 0.4579 | 0.1984 | 0.1801 | 0.5464 | 0.4810 |
| (C) | CNN-rand | 0.1653 | 0.0896 | 0.5599 | 0.5361 | 0.2941 | 0.2862 | 0.6740 | 0.6790 |
| | CNN-pre | 0.3658 | 0.3752 | 0.5870 | 0.5722 | 0.6373 | 0.6355 | 0.6775 | 0.6774 |
| | LSTM-rand | 0.1987 | 0.1409 | 0.5417 | 0.5396 | 0.1284 | 0.0855 | 0.6333 | 0.5939 |
| | LSTM-pre | 0.2806 | 0.2781 | 0.5462 | 0.4411 | 0.4841 | 0.5059 | 0.6446 | 0.6443 |
| | TLGNN | 0.4933 | 0.4997 | 0.5500 | 0.6148 | 0.6937 | 0.6079 | 0.6262 | 0.6158 |
| | TextING | 0.4364 | 0.4471 | 0.5684 | 0.5631 | 0.6520 | 0.6521 | 0.6622 | 0.6605 |
| | HyperGAT | <u>0.6146</u> | <u>0.6123</u> | 0.5568 | 0.4944 | 0.6475 | 0.6478 | 0.7217 | 0.7184 |
| | TextGCN | 0.5343 | 0.5247 | 0.6042 | 0.5940 | 0.6254 | 0.6177 | 0.7312 | 0.7258 |
| | TensorGCN | 0.2425 | 0.1941 | 0.5203 | 0.4980 | 0.3726 | 0.3682 | 0.5449 | 0.4534 |
| (D) | STCKA | 0.5440 | 0.5345 | 0.6412 | 0.6394 | 0.5354 | 0.5363 | 0.6416 | 0.6391 |
| | HGAT | 0.5711 | 0.5663 | 0.6188 | 0.6178 | 0.5211 | 0.5137 | 0.6556 | 0.6546 |
| | SHINE | 0.5118 | 0.5074 | 0.5817 | 0.5811 | 0.6681 | 0.6451 | 0.6441 | 0.6440 |
| | GIFT | 0.5343 | 0.5111 | <u>0.7372</u> | <u>0.7372</u> | <u>0.7211</u> | <u>0.6969</u> | 0.7976 | <u>0.7975</u> |
| | **LIGRAM** | **0.8403** | **0.8269** | **0.7449** | **0.7447** | **0.8049** | **0.7986** | **0.8128** | **0.8115** |
| (E) | GPT-5.2 | 0.5146 | 0.5197 | **0.7899** | **0.7899** | 0.6004 | 0.5983 | **0.9067** | **0.9067** |
| | Gemini-2.0-flash | 0.5147 | 0.4544 | 0.7364 | 0.7266 | 0.5894 | 0.5800 | 0.8782 | 0.8778 |
| | Qwen2.5-7B | 0.2671 | 0.2454 | **0.7469** | **0.7462** | 0.2468 | 0.2015 | 0.8159 | 0.8127 |
| | Qwen3-4B | 0.6022 | 0.5622 | **0.9032** | **0.9032** | 0.2321 | 0.1896 | 0.3394 | 0.3280 |

### 4.3  Implementation Details

The proposed model is implemented in PyTorch and conducted on an NVIDIA GeForce RTX 4090 GPU. Each linguistic subgraph (morpheme, part-of-speech, and named entity) is encoded using a two-layer GCN with a hidden dimension of 200. The key hyperparameters were empirically determined through iterative performance tuning across various combinations to identify the optimal configuration for Korean text. For PMI, the sliding window size is set to 5 after evaluating multiple candidates to maximize local context capture while minimizing noise, and the similarity threshold ($\delta$) is set to 2.7 to ensure only semantically significant relations are represented. The dropout rate is 0.7. The weighting coefficient ($\lambda$) for the contrastive loss is set to 0.7, balancing the contribution of the classification and contrastive objectives. We employ the AdamW optimizer with a learning rate of $5 \times 10^{-4}$ and a weight decay of $1 \times 10^{-3}$. The model is trained for up to 1000 epochs, with the best validation performance evaluated every five epochs and saved for the test set. This setup reflects a semi-supervised scenario consistent with standard STC tasks.

### 4.4    Results

Table 2 reports the performance of LIGRAM and baseline models on four Korean short-text datasets (KLUE YNAT, Movie Reviews, Snippets, Shopping) using Accuracy (ACC) and Macro F1-score (F1).

Graph-based models generally outperform traditional and simple neural baselines (e.g., CNN, LSTM). Among all methods, LIGRAM achieves the best results across datasets. On YNAT, LIGRAM reaches 0.8403/0.8269 (ACC/F1), achieving a +21.5% F1 improvement over HyperGAT. On Snippets, it attains 0.8049/0.7986, yielding a +8.4% ACC improvement over GIFT. It also records the highest scores on Movie Reviews (0.7449/0.7447) and Shopping (0.8128/0.8115), indicating robust cross-domain generalization.

These gains stem from explicitly modeling Korean linguistic properties (an agglutinative structure, flexible word order, and frequent particle omission) through morpheme ($w$), POS ($p$), and entity ($e$) subgraphs that are hierarchically integrated to refine document representations in short contexts. Semantics-aware contrastive learning further improves topic-level discrimination by aligning semantically similar documents. Together, linguistically informed graph modeling and SemCon enable accurate and semantically consistent classification even under label-scarce conditions, demonstrating the effectiveness and scalability of LIGRAM for low-resource Korean STC.

### 4.5    Comparison with Large Language Models

On some datasets (MR and Shopping), LLMs achieved higher performance; for example, GPT-5.2 and Qwen3-4B reach around 0.79–0.90, surpassing our model on these binary sentiment-oriented tasks. These datasets rely heavily on surface lexical cues and sentiment expressions, and LLMs, pre-trained on large-scale corpora, are exposed to diverse sentiment patterns, which aligns with their strong performance. For closed-source models such as GPT or Gemini, pre-training data composition is not publicly available, limiting analysis of data source effects.

Notably, relatively small LLMs (e.g., 4B-scale) also show strong results, suggesting that performance does not strictly scale with model size but may relate to pre-training data and learned linguistic patterns. Although LLMs were fine-tuned on the same training data, datasets with many classes require fine-grained category distinctions where discriminative lexical cues become more critical. Consistent with this, our model achieves the best results on multi-class datasets such as YNAT (0.8403/0.8269) and Snippets (0.8049/0.7986 in ACC/F1).

LLMs are mainly optimized for next-token prediction in text generation, whereas short-text classification depends on discriminative sentence representations. Under these different objectives, representation-focused models remain competitive. Moreover, despite using only 0.56M parameters, the proposed model maintains competitive performance across diverse short-text settings compared to large-scale LLMs.

**Table 3.** Ablation Study

| Model | KLUE YNAT | | Movie Reviews | | Snippets | | Shopping | |
|---|---|---|---|---|---|---|---|---|
| | ACC | F1 | ACC | F1 | ACC | F1 | ACC | F1 |
| morpheme | 0.6589 | 0.6496 | 0.5393 | 0.4524 | 0.7430 | 0.7317 | 0.6282 | 0.5885 |
| pos | 0.2114 | 0.1918 | 0.5356 | 0.5406 | 0.1856 | 0.1308 | 0.5292 | 0.5256 |
| entity | 0.1429 | 0.0357 | 0.4895 | 0.3286 | 0.2197 | 0.0450 | 0.4962 | 0.3316 |
| morpheme/pos | 0.6166 | 0.6103 | 0.5627 | 0.5373 | 0.7230 | 0.7163 | 0.7162 | 0.7108 |
| morpheme/entity | 0.6767 | 0.6616 | 0.6123 | 0.6085 | 0.7503 | 0.7445 | 0.6720 | 0.6528 |
| pos/entity | 0.2265 | 0.2173 | 0.5264 | 0.5012 | 0.1879 | 0.1304 | 0.5403 | 0.5385 |
| w/o SemCon | 0.6365 | 0.6161 | 0.6796 | 0.6762 | 0.7691 | 0.7557 | 0.7437 | 0.7418 |
| **LIGRAM** | **0.8403** | **0.8269** | **0.7449** | **0.7447** | **0.8049** | **0.7986** | **0.8128** | **0.8115** |

### 4.6 Ablation study

To quantify the contribution of each component in LIGRAM, we conducted an ablation study under identical settings using Accuracy and Macro F1-score.
Table 3 reports results for different subgraph combinations and the inclusion of Sem-Con. With a single graph, the morpheme graph showed stable performance by capturing lexical semantics, whereas POS-only and entity-only graphs performed substantially worse, indicating that syntactic or entity-level cues alone are insufficient to recover the overall semantic context of short texts. When two cues were combined, configurations including the morpheme graph (morpheme/POS, morpheme/entity) consistently outperformed others, and the morpheme/entity combination yielded strong performance across datasets by effectively integrating contextual meaning with core entity information.

The model without SemCon preserved linguistic integration but lacked semantic alignment across documents, resulting in reduced discriminability between semantically related classes. In contrast, the full LIGRAM achieved consistent gains on all datasets, including +21.1% F1 on YNAT and +7.8% F1 on Movie Reviews. Overall, SemCon provides an average F1 improvement of 9.8%, confirming that semantics-aware contrastive learning effectively aligns semantically similar documents and clarifies inter-class semantic boundaries.

These results show that LIGRAM improves short-text classification by (1) hierarchically integrating morpheme-level, POS-level, and entity-level graphs to capture fine-grained grammatical and semantic dependencies and (2) applying SemCon to refine the representation space, thereby enhancing semantic boundary discrimination and generalization under limited context.

## 5      Conclusion

In this study, we proposed LIGRAM, a graph-based model designed for Korean short-text classification. By constructing subgraphs at the morpheme, part-of-speech, and named entity levels and hierarchically integrating them, the model effectively captures the agglutinative and syntactic characteristics of the Korean language. In addition, Sem-Con enables the model to incorporate semantic similarity among documents and achieve clearer topic separation even in cases with ambiguous topic boundaries. Experimental results on four datasets demonstrate that LIGRAM consistently outperforms all baseline models.

These findings confirm that the combination of linguistically informed graph modeling and SemCon enables effective representation learning in short-text environments.

As future work, we plan to explore how the proposed linguistically informed graph framework can be adapted to other languages and evaluate its cross-lingual generalization capability.

**Disclosure of Interests.** The authors have no competing interests to declare that are relevant to the content of this article.

**Acknowledgments.** This work was supported by (i) National Research Foundation of Korea (NRF) grants funded by the Korea government (MSIT) (RS-2021-NR061821, RS-2023-00242528), (ii) Institute of Information & Communications Technology Planning & Evaluation (IITP) programs for innovative human-resource development, software excellence, and local intellectualization funded by MSIT (IITP-2025-RS-2023-00260267, 2025-0-00058), and (iii) the 2025 Research Grant from Kangwon National University. The corresponding authors are Yeongwook Yang and Hong-Jun Jang.